\documentclass[letterpaper, 10 pt, conference]{ieeeconf}  % Comment this line out if you need a4paper

\IEEEoverridecommandlockouts                              % This command is only needed if 
\usepackage{graphics} % for pdf, bitmapped graphics files
\usepackage{epsfig} % for postscript graphics files
\usepackage{mathptmx} % assumes new font selection scheme installed
\usepackage{times} % assumes new font selection scheme installed
\usepackage{amsmath} % assumes amsmath package installed
\usepackage{amssymb}  % assumes amsmath package installed
\usepackage[font=small]{caption}
\usepackage{paralist}
\usepackage{threeparttable}
\usepackage{booktabs}
\usepackage{stix}
\makeatletter
\let\NAT@parse\undefined
\makeatother
\usepackage{hyperref} 
\usepackage{multirow}
\usepackage{amssymb}
\newcommand{\etal}{\textit{et al.}~}
\newcommand{\nCr}[2]{C^{#2}_{#1}} % nCr
\title{\LARGE \bf
Robotic Interestingness via Human-Informed Few-Shot Object Detection}

\author{Seungchan Kim, Chen Wang, Bowen Li, Sebastian Scherer% <-this % stops a space
\thanks{The authors are with the Robotics Institute, Carnegie Mellon University, Pittsburgh, PA 15213, USA. {\tt\small seungch2@andrew.cmu.edu; chenwang@dr.com; bowenli1024@gmail.com; basti@andrew.cmu.edu}}%
}

\newcommand{\fref}[1]{Fig.~\ref{#1}}
\newcommand{\sref}[1]{Section~\ref{#1}}
\newcommand{\tref}[1]{Table~\ref{#1}}

\newcommand{\ie}{{i.e.},~}
\newcommand{\eg}{{e.g.},~}

\begin{document}

\maketitle
\thispagestyle{empty}
\pagestyle{empty}

%%%%%%%%%%%%%%%%%%%%%%%%%%%%%%%%%%%%%%%%%%%%%%%%%%%%%%%%%%%%%%%%%%%%%%%%%%%%%%%%
\begin{abstract}
Interestingness recognition is crucial for decision making in autonomous exploration for mobile robots. Previous methods proposed an unsupervised online learning approach that can adapt to environments and detect interesting scenes quickly, but lack the ability to adapt to human-informed interesting objects. To solve this problem, we introduce a human-interactive framework, AirInteraction, that can detect human-informed objects via few-shot online learning.
To reduce the communication bandwidth, we first apply an online unsupervised learning algorithm on the unmanned vehicle for interestingness recognition and then only send the potential interesting scenes to a base-station for human inspection.
The human operator is able to draw and provide bounding box annotations for particular interesting objects, which are sent back to the robot to detect similar objects via few-shot learning. Only using few human-labeled examples, the robot can learn novel interesting object categories during the mission and detect interesting scenes that contain the objects. We evaluate our method on various interesting scene recognition datasets. To the best of our knowledge, it is the first human-informed few-shot object detection framework for autonomous exploration.
\end{abstract}

%%%%%%%%%%%%%%%%%%%%%%%%%%%%%%%%%%%%%%%%%%%%%%%%%%%%%%%%%%%%%%%%%%%%%%%%%%%%%%%%
\section{Introduction}
\begin{figure}[!t]
    \centering
    \includegraphics[width=1.0\linewidth]{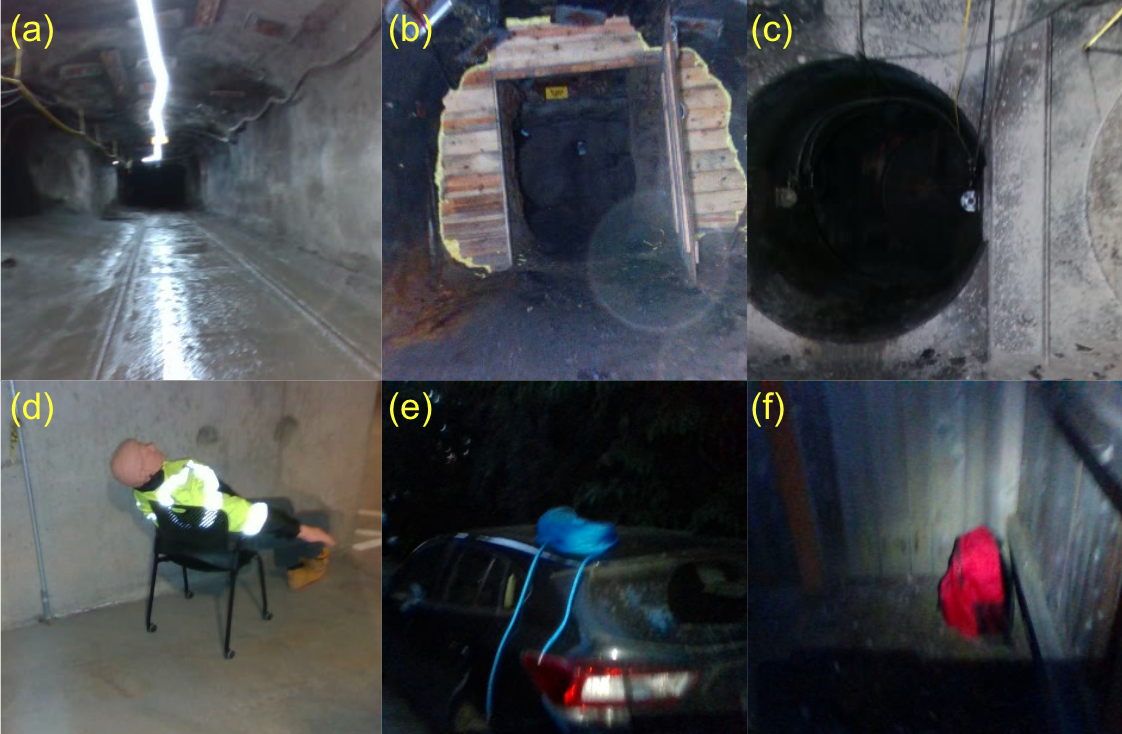}
    \caption{Exemplary images from robots' exploration in the DARPA Subterranean Challenge. Some scenes, \eg (a) are uninteresting, while there are interesting scenes \eg (b), (c), (d), (e), and (f).}
    \label{intro-figure}
\end{figure}

Interestingness recognition is a crucial component of autonomous mobile robot exploration. Robots should perceive the surroundings by processing visual information quickly and estimate the interestingness of the scenes to make decisions for next actions and plans. Take a robot that explores underground tunnel as an example (first row of \fref{intro-figure}). While an uninteresting scene like \fref{intro-figure} (a) may not change the course of the robot actions, interesting scenes that contain the wooden door as in \fref{intro-figure} (b) or a dark entrance hole as in \fref{intro-figure} (c) would affect the next actions of the robots, such as moving towards the door or hole.

There have been various approaches to modeling visual interestingness using the concepts of saliency \cite{borji-et-al, learning-saliency-based-attention}, memorability \cite{photo-memorable}, and novelty detection \cite{novelty-detection}. However, these previous algorithms were not apt for robotic applications, because they were unable to adapt to unknown environments quickly and perform real-time responses to the rapidly changing scenes. Recently, Wang \etal proposed an unsupervised learning approach that detects visually interesting scenes in an \textit{online} manner \cite{eccv, wang2021unsupervised}. The core of this algorithm is the use of a visual memory module that adapts to the surrounding environment quickly and remembers input scenes efficiently, so that the robot can recall the past and detect novel, unusual scenes as interesting in real-time \cite{wang2021unsupervised}. Another notable feature is that, even if the robot once detects a specific scene as interesting, it loses interest over following similar scenes gradually if they appear repetitively, just like humans lose interest when exposed with a series of similar scenes.

However, one limitation of this algorithm is that it does not have the capability to adapt to human-informed interesting objects. Take this scenario as an example: an unmanned aerial vehicle (UAV) is exploring an underground tunnel and encounters scenes that look interesting and contain different objects as shown in \fref{intro-figure} (d), (e), and (f). The previous unsupervised online learning algorithm \cite{eccv} enables the UAV to detect these scenes as interesting, but it can't specify which objects within the scenes make them actually interesting. Depending on the missions and human operator's interests, the targeted interesting objects might differ with respect to the context. To solve this problem, we aim to take human operators' feedback into account for better recognition of interesting scenes and objects. 

\begin{figure*}[!t]
    \centering
    \includegraphics[width=1.0\textwidth]{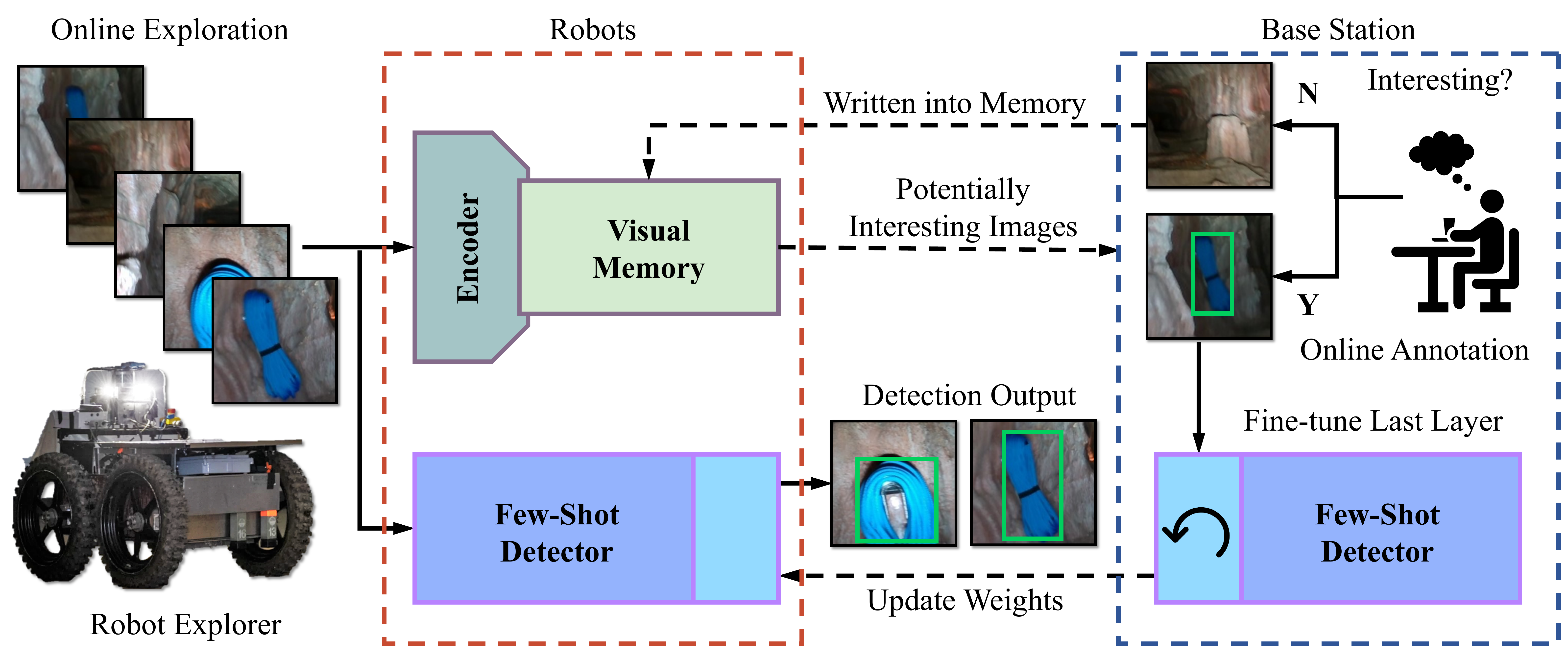}
    \caption{Diagram of AirInteraction. The robot explorer perceives continuous streams of images, which first go through the unsupervised visual interestingness module. The interesting scenes detected by the visual memory are sent to the base station, where the human annotator can decide if the image is interesting or not. Uninteresting images are sent back to the robot and written into the memory, which prevents the false judgement of similar scenes afterwards. The interesting objects in the real interesting images are annotated by the human (drawing tool is given). Then the last layer of the few-shot object detector is fine-tuned on the novel object using the provided annotations. The updated weights of the final layer are copied to the classifier onboard periodically, so that the robot can detect novel interesting objects in real-time during the mission.}
    \label{fig:diagram}
\end{figure*}

The challenge is that, before a mission starts, \textit{both} the robot and human operator may not have information about what the interesting objects would be. During the exploration, the robot may discover a completely novel object, which is possibly uncommon for existing large scale training set \cite{MSCOCO} and even could belong to an unknown category to the human operator. However, it could still be interesting to the operator, and there is a need for the robot to learn about the object quickly \textit{during the mission} and detect similar objects in the later phase of the exploration. Hence, traditional object detection approaches \cite{faster-rcnn,yolo,ssd} trained for fixed and predefined categories on existing large scale datasets \cite{MSCOCO} are inapplicable to this situation. 

We suppose that human's few online inputs on the object-specificity, (\eg certain objects or specific image parts that make the scene interesting), can be of great assistance to robot exploration.
Specifically, when a robot detects an interesting scene (image) online, it sends the potential interesting image to a human operator at the base station.
% who are able to provide essential feedback to the few-shot object detection algorithms. 
When the human operator thinks an image is interesting, the operator is prompted to draw a bounding box around the interesting object/part within the image.
This image and bounding box annotation are fed into a few-shot object detector \cite{TFA}, which can quickly learn an object detection model only using a very few labeled examples. Since human annotations can be expensive, we restrict the model to learn novel, unseen objects using just one, two, or three examples per category.

In summary, our contributions are:
\begin{itemize}
\item To solve the problem of existing algorithm that can only detect predefined object categories, we propose a human-interactive framework, AirInteraction, for autonomous robots to detect visually interesting and human-informed objects during online exploration.
\item To reduce human efforts and communication bandwidth, we design three components: (1) an online unsupervised algorithm running on the exploration robot to detect visually interesting scenes, (2) a human-robot interaction interface running on a base-station to process operator's feedback for interesting objects, and (3) a few-shot object detector running across the robot and the base-station to enable fast training of human-informed objects during the robotic mission.
% \item Building upon a visual memory module and online adaption method, we enable the robots to request and get feedback from human operator on the interestingness of the scenes. Our method enables robots to learn about novel, unseen objects \textit{during the mission} with the minimal amount of help from human operator. To this end, we integrate few-shot object detection module with the online interesting scene recognition algorithm. Dynamically adding labelled image-annotation examples, few-shot object detector trains its own predictor, and helps robots to detect more relevant, interesting scenes.
\item We test the effectiveness of AirInteraction on various challenging datasets and perform ablation studies to verify the necessity of each component.
\end{itemize}

In \sref{sec:background}, we review prior works on robotic interestingness and few-shot object detection. Our new method, human-in-the-loop online visual interestingness with few-shot object detection is presented in \sref{sec:method}.
The experimental setup, results of our method and baselines, and the ablation study are shown in \sref{sec:experiment}, followed by the limitation of our framework and possible solutions in \sref{sec:limitation}.

\section{Related Work}\label{sec:background}

\subsection{Visual Interestingness}

Although prior works \cite{what-is-interesting}\cite{interestingness-survey} have called the extent to which an image/video captivates or holds one's attention as visual interestingness, the strict definition and quantification of visual interestingness hasn't been globally agreed yet. Since the notion of interestingness is inherently subjective and non-trivial \cite{interestingness-survey}, there have been many different approaches to model visual interestingness using various concepts. Isola \etal \cite{photo-memorable} and Bylinskii \etal \cite{bylinskii} focused on the concept of  memorability, investigating the intrinsic and extrinsic factors that make images remembered by humans' memory. Borji \etal \cite{borji-et-al} and Zhao \etal \cite{learning-saliency-based-attention} focused on the concept of saliency, investigating features of subparts, objects, or regions within an image that stand out from neighboring parts, and reviewed saliency-based visual attention that drives the shift and focus of human gaze. Gygli \etal \cite{Gygli2013} proposed computational evidence for different visual cues that constitute interestingness of images, especially focusing on aesthetics, unusualness, and preferences. Hsieh \etal \cite{social-interesting} examined the interaction between visual and social interestingness, focusing on virality, popularity, and aesthetics features. However, these aforementioned works have focused on \textit{static} properties of images, which is not suitable for situations that require adaptation to the changing distributions.
Previous works \cite{eccv,wang2021unsupervised} proposed a new concept of visual interestingness for robot exploration with online adaptation to changing distributions.
It can report the potentially interesting frames, but fail to detect user-informed objects in real-time. In this paper, we will solve this problem by incorporating a few-shot object detection model.

\subsection{Few-Shot Object Detection}
Object detection aims to estimate the locations and predefined categories of objects in an image. With the emergence of deep learning-based methods, there have been rapid progress of object detection algorithms, mainly divided into two branches of algorithms: (1) two-stage approaches \cite{fast-rcnn,faster-rcnn,R-FCN,mask-rcnn} that separate region proposal generation and proposal classification/regression, (2) one-stage approaches \cite{yolo,ssd} that design an end-to-end neural network to tackle the problem. However, the limitation of the general object detection is the requirement of the large scale annotated training data. Moreover, the object categories are fixed and predefined after training.

Along with the recent progress in few-shot learning \cite{vinyals2016,maml} that learns to generalize with only a few novel data, there have emerged various approaches to tackling few-shot object detection task. A few-shot detector aims to detect novel object classes using merely a few labeled examples. One paradigm for few-shot object detection is transfer learning \cite{TFA, LSTD, mpsr}, which pretrains a base model on base classes using abundant data, and adapts the model via fine-tuning on few labeled examples. Another paradigm is meta-learning \cite{metarcnn, feature-reweighting, ONCE, meta-detect-rare,airdet}, which aims to extract meta-level knowledge during base training that enables efficient learning on novel classes with a few examples.

\section{Methodology} \label{sec:method}

\subsection{Overview}

As shown in \fref{fig:diagram}, AirInteraction is composed of three components, including (1) an online unsupervised learning algorithm running on the robot that detects interesting scenes during exploration in real-time, (2) a human-robot interactive interface at the base station, and (3) a few-shot object detector that learns human-informed objects online.

The workflow of AirInteraction is also simple but efficient. Due to the uncertainty of working environments, we cannot guarantee the network condition, thus we reduce the communication bandwidth requirement by putting most of the computation on the robot and only send necessary information to the base-station.
Specifically, the robot detects potentially interesting scenes by employing the online interestingness detection algorithm and only sends the images with highest probability to be interesting to the base-station.

If the network is interrupted, the interesting scenes will be stored in a stack and cover those images with lowest probability. The images with higher probability will be sent in higher priority once the network is established again.
% If the human operator with the base-station thinks one image received is not interesting, the image ID is sent back to the robot to update the interestingness detection module so that similar images will not be sent again, otherwise 
If the human operator with the base-station thinks an image is interesting, the operator will be prompted to select the region of interests, such as a novel object. The selected regions (bounding boxes) and the object labels are then taken as training samples to fine-tune the few-shot object detector.

The few-shot detector is fine-tuned on the base-station and a mirror copy model runs on the robot.
The human operator's annotations are dynamically added to the training samples' pool to update the model parameters, which are sent to the robot periodically to detect similar objects.
We next explain more details about the three modules, respectively.

\begin{figure}[t]
    \centering
    \includegraphics[width=1\linewidth]{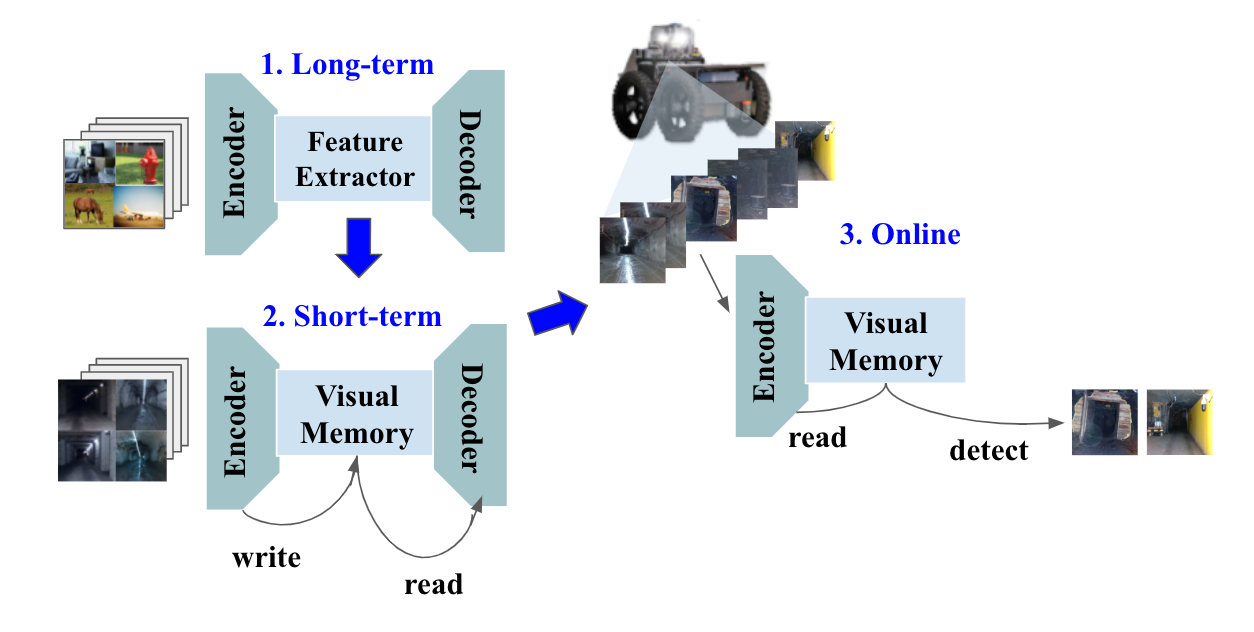}
    \caption{The structure of the unsupervised online learning model. It detects the potentially interesting scenes based on a visual memory module. The images that are not well recalled by the online updated memory are taken as interesting, which will be sent to base-station.}
    \label{fig:online-learning}
\end{figure}

\subsection{Online Unsupervised Interestingness Detection}

To reduce the communication requirement, we leverage the unsupervised online learning model \cite{eccv,wang2021unsupervised} to select potentially interesting frames.
This approach is composed of a three-stage learning architecture, i.e., (1) the \textit{long-term} learning, which acquires general world-level knowledge on large scale training data, (2) the \textit{short-term} learning, that learns prior task-dependent knowledge via a visual memory module from hundreds of uninteresting images in several minutes, and (3) the online learning, which processes image stream and updates the memory module in real-time to adapt to the environments.

The key of this architecture is the visual memory, which saves the knowledge about the uninteresting scenes during the short-term learning, and outputs lower confidence when novel scene appears suddenly during the online mission execution. One important feature of this system is a losing of interests over repetitive scenes, \ie when a robot sees an interesting image, it is quickly written into the memory, so that the interestingness level of similar following scenes decrease gradually. However, the interestingness recognition system \cite{eccv,wang2021unsupervised} has no interaction with the operators and it cannot detect specific human-informed objects.

When the exploration robot perceives the surrounding environment, it processes continuous streams of image. Each image vector $\textbf{x}(t)\in \mathbb{R}^{C\times W \times H} $ goes through the visual memory module $\textbf{M} \in \mathbb{R}^{N \times C \times W \times H}$ of the interestingness algorithm, where $C,W,H$ are number of channels, width, and height of the image, and $N$ is the number of cubes in the visual memory. When the robot sees a new scene $\textbf{x}(t)$ at time $t$, the visual memory module $\textbf{M}$ is updated as follows:
\begin{equation}
    \textbf{M}_i(t) = (1-w_i)\cdot \textbf{M}_i(t-1) + w_i\cdot \textbf{x}(t),
\end{equation}
where $w_i$ is the $i$-th element of the vector $\textbf{w}\in\mathbb{R}^{N}$
\begin{equation}
\textbf{w} = \sigma\Big(\gamma_w \cdot \tan\big(\frac{\pi}{2} \cdot \mathcal{D}\big(\textbf{x}(t),\textbf{M}(t-1)\big)\big)\Big),
\end{equation} 
where $\sigma$ is the {\tt softmax} function, $\gamma_w$ is a learning speed control parameter, and $\mathcal{D}$ is the cosine similarity function. This \textit{writing} process \cite{eccv, wang2021unsupervised} enables the visual memory module to incrementally learn new scenes without forgetting old scenes.

Concurrently, when the robot sees a scene $\textbf{x}(t)$, it needs to recall the memory and compare the new scene with the old scenes within the memory. Within this memory $reading$ process \cite{eccv, wang2021unsupervised}, the reading score is defined as
\begin{equation}
 f(t) = \sum_{i=1}^{N} v_i \cdot \textbf{M}_i(t),
\end{equation}
where $v_i$ is the $i$-th element of the vector $\mathbf{v}\in\mathbb{R}^{N}$
\begin{equation}
    \mathbf{v} = \sigma\Big(\gamma_v \cdot \tan\big(\frac{\pi}{2} \cdot \mathcal{S}\big(\textbf{x}(t), \textbf{M}(t)\big)\big)\Big),
\end{equation} 
where $\gamma_v$ is also a learning speed control parameter for reading, and $\mathcal{S}$ is maximum cosine similarity operator between $\textbf{x}(t)$ and all possible translations of $\textbf{M}(t)$, which is calculated from element-wise multiplication of their Fourier transforms.

Intuitively, when the memory reading confidence score is low, the scene is more likely to be interesting since the scene has never been visited before, and vice versa. In practice, we take the cosine similarity between $f(t)$ and $x(t)$ as the reading confidence.
All the scenes that were previously seen are written into the memory; this prevents robots over-detecting similar scenes as interesting. Even if a scene is interesting at first glance, robot still loses interests over the scene, if the scene is repetitively visited.

We choose the visual memory learning scheme \cite{eccv, wang2021unsupervised} as the first component of AirInteraction, not only due to the high efficiency of the algorithm, which enables real-time online processing of image streams, but also because it can make prior judgments so that only a few potentially interesting scenes will be sent to the base station. Using the visual memory, the robot can respond to novel scenes only, ignore the repetitive or similar scenes, and avoid the redundant request for human feedback.

\begin{figure}[!t]
    \centering
    \includegraphics[width=1.0\linewidth]{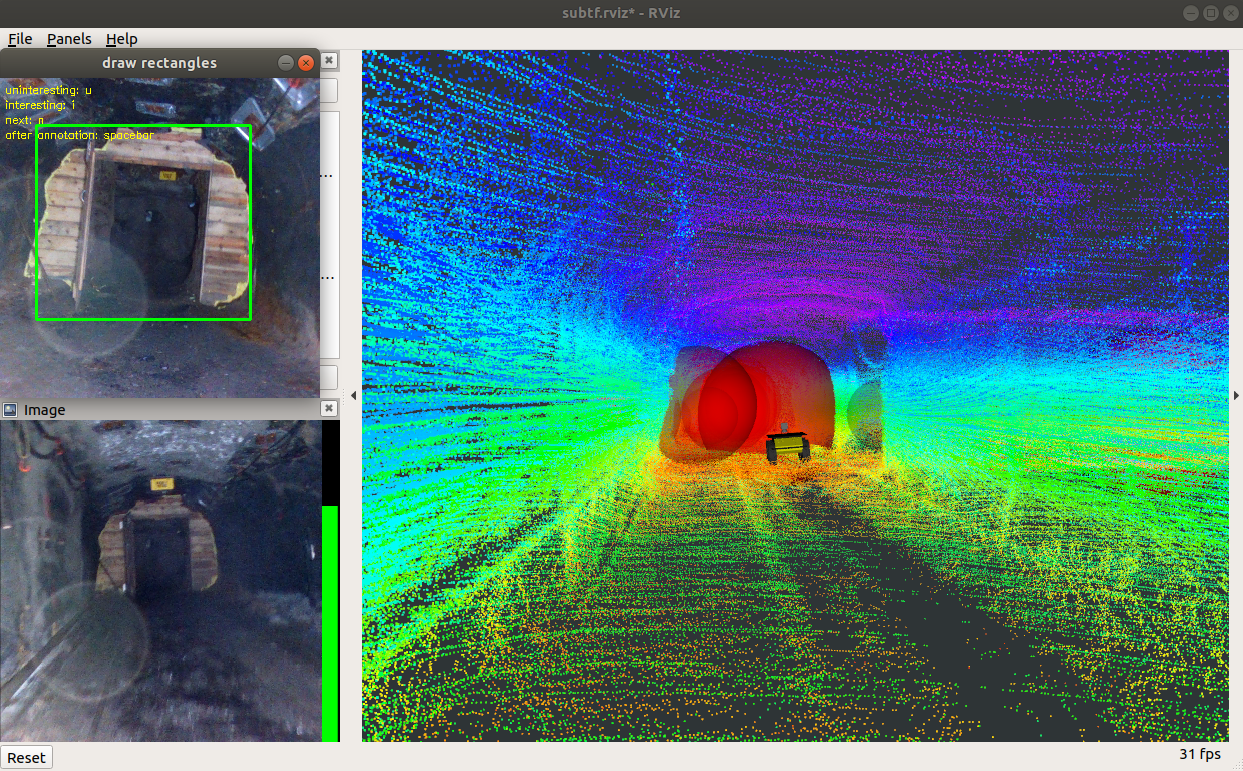}
    \caption{Human-robot interface for annotation and surveillance.}
    \label{fig:interface}
\end{figure}

\subsection{Human-Robot Interactive Interface}
We design the interface for human-robot interaction between the robot explorer and human operator based on OpenCV \cite{culjak2012brief} and ROS \cite{quigley2009ros}.
A screenshot of the visualization and annotation tool is shown in the \fref{fig:interface}. 
When the interesting scenes detected by the robot are sent to the base station, the scenes are queued and being shown to the human operator one by one. If the human operator presses the key {\tt N}, which denotes \textit{not interesting}, the image is sent back to the robot and written into the memory, which prevents similar following scenes from being detected as interesting by the robot. 

If the human operator presses the key {\tt I}, which means the human operator thinks the scene is \textit{interesting}, the pop-up screen of the image appears where the operator can draw bounding box of the interesting object in the image. The annotation information of the image and the location of the bounding box are sent to the few-shot object detector running on the base station, which will be explained next subsection.

Our interface is designed to make the human operator's interaction with robot as easy and simple as possible.
Using this visualization tool, any inexperienced human operator can give feedback to the robot with minimal efforts and actions by just clicking the image screen and using a mouse to draw the bounding box.

\subsection{Few-shot Object Detector}
The human-informed object annotations and the image pairs serve as few labeled example inputs for few-shot object detector, which is equipped in the base station. We further explain the details about the choice of the few-shot object detection model and techniques to improve its performance.
\subsubsection{Two-stage Fine-tuning Approach}
In this work, we employ one of the transfer-learning based few-shot object detection model, two-stage fine-tuning approach (TFA) \cite{TFA} as a baseline, which is known to have state-of-the-art performance on benchmarks. The two-stage refers to (1) the base training stage which jointly trains feature extractor and box predictor for base classes, and (2) the fine-tuning stage which trains box predictor for novel classes. For the base model, we follow the same architectures proposed by \cite{TFA}, where Faster R-CNN \cite{faster-rcnn} is adopted as a base detector with Resnet-101 \cite{deep-residual} backbone. 

During the fine-tuning stage, the feature extractor of the model is fixed, and only the final layer of the box predictor is fine-tuned. While training for base classes requires abundant data, training for novel classes via fine-tuning the final layer only requires using few labeled examples. We use the human-informed image-annotation as the labeled examples and feed them into the trainer to fine-tune the final layer of the model.

\subsubsection{Weighted Mixture of Batch for Fine-tuning}
\label{sec:weighted}

When fine-tuning the last layer of the model, previous work \cite{TFA} used the balanced mixture of base class images and novel class images as a minibatch for training. However, our experiments show that this often cannot produce satisfactory results for novel classes. Instead, AirInteraction employs a weighted mixture of batch for fine-tuning. Let's denote the number of base classes and novel classes as  $B$ and $N$, and we use $K$ instances per category during fine-tuning. The random mixture of $BK$ base images and $NK$ novel images are used. If the minibatch size is $m$, each minibatch is a random combination of base and novel images, and there are $\nCr{(B+N)K}{m}$ combinations. In this work, we use a variant of this scheme by increasing the frequency of using novel (human-informed) images when generating a minibatch, rather than using random uniform mixture. Let's denote the ratio of novel images compared to base images as $r~(r\ge 1)$. Then there can be $\nCr{(B+rN)K}{m}$ combinations of minibatches to be used for fine-tuning. Note that we are not changing the number of instances per novel category, but just simply reusing the existing instances more frequently to generate minibatches. 
\begin{figure}[!t]
    \centering
    \includegraphics[width=0.8\linewidth]{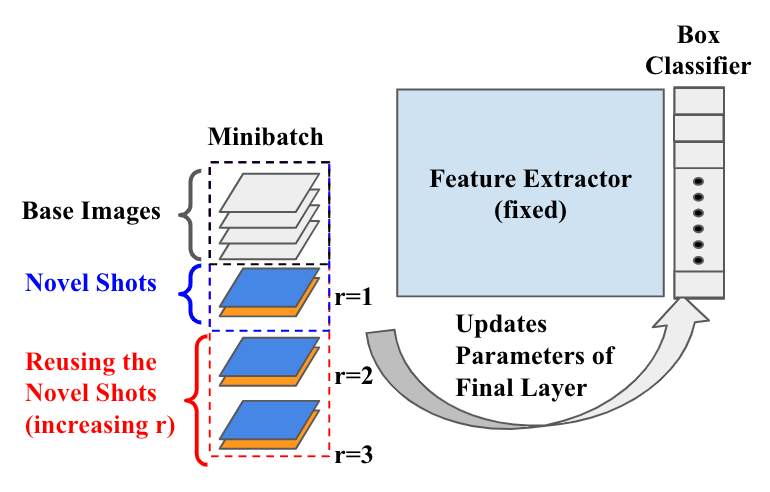}
    \caption{The combination of base images and novel image shots compose the batch for the fine-tuning training. We parameterize the frequency of using novel image shots compared to base images, and use the novel shots up to 3 times more frequently than base images.}
    \label{fig:fine-tuning}
\end{figure}

\subsubsection{Parameter synchronization between robot and base station} 
We have two identical few-shot object detectors in the system, one in the base station, and the other in the robot. Since training the object detection model is computationally heavy for the robot, we perform the fine-tuning process in the base station. Note that the parameters of the pretrained features are fixed, both in the robot and the base station, and only the parameters of the final layer are updated. Therefore, it suffices to synchronize the final layer parameters of the object detector in the base station and the one in the robot. The dimension of the final layer is comparatively small, and can be sent easily to the robot for synchronization.

\section{Experiments \& Results}\label{sec:experiment}

\begin{figure}[!t]
    \centering
    \includegraphics[width=0.85\linewidth]{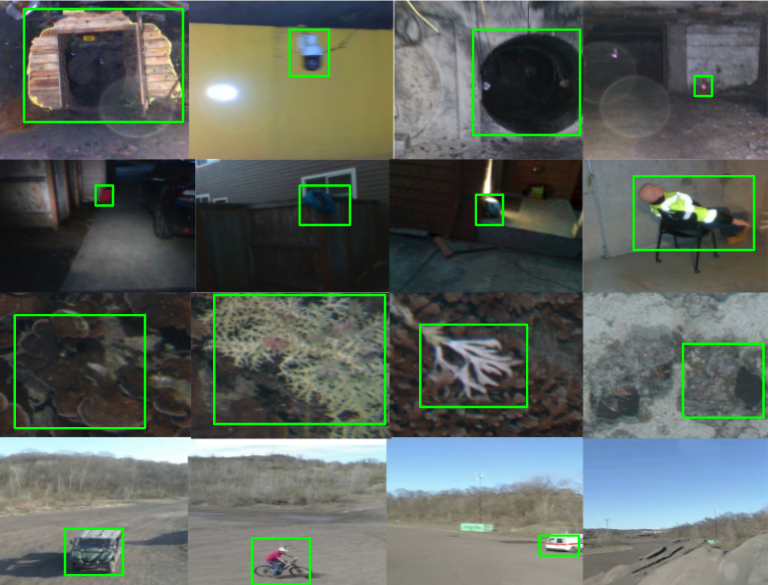}
    \caption{Representative images from each dataset. From 1st to 4th row: SubT Tunnel, SubT Urban, Scott Reef 25, Drone Filming}
    \label{fig:dataset-images}
\end{figure}

\subsection{Dataset}
The experiments are conducted on four authoritative challenging datasets, as shown in \fref{fig:dataset-images}, to verify the effectiveness of our system.

\subsubsection{SubT Tunnel and SubT Urban Datasets}
The two datasets are recorded during the DARPA Subterranean (SubT) Challenge, which requires competitors to build autonomous robotic systems to search and explore the subterranean environments like tunnel and urban underground. In this challenge, the autonomous robots are prompted to search for different objects. We use the datasets recorded by two autonomous UGVs from the Team Explorer (Carnegie Mellon University and Oregon State University).
For SubT Tunnel circuit, we chose door, hole, surveillance camera, and fire extinguisher as interesting objects, and for SubT Urban dataset, we chose helmet, backpack, rope, drill, vent, and human survivor as interesting objects.

\subsubsection{Scott Reef 25 Dataset} The Scott Reef 25 dataset is a sequence of top-down view images recorded by autonomous underwater vehicle (AUV) at Scott Reef in Western Australia. 
% Among the class labels in this dataset (sand, finger coral, mushroom coral, weed, and debris), 
There are five labeled objects in this dataset, in which we take sand as uninteresting and the other objects including finger coral, mushroom coral, weed, and debris as interesting.
% We split the objects into two categories: sand as uninteresting, and other 4 objects (finger coral, mushroom coral, weed, and debris) as interesting. 

\subsubsection{Drone Filming Dataset} It is an outdoor dataset recorded by quadcopters that follow moving targets during real-time aerial filming. In this work, we select three different object categories (van, truck, bicycler) as interesting.

\subsection{Baseline, Metrics, and Hardware}

% \subsubsection{Baseline}
Since AirInteraction is the first human-informed robotic interestingness detection system, we lack off-the-shelf systems to compare.
Therefore, we will take the few-shot object detection method TFA \cite{TFA} as a baseline. For fairness, we adopted the default parameters for the TFA. To test the performance of few-shot object detector trained on the inputs from the human operator, we use the traditional evaluation metric, Average Precision (AP) score \cite{MSCOCO} for each class of objects and across the classes. For each dataset, we measure and average the performances over the object classes. For fine-tuning, we used 1 Nvidia GeForce RTX 2060 GPU. 
% trained with different human inputs. 

% We measure the performance of the object detector of the AirInteraction trained only using few labeled examples provided by human annotator during the mission. 
% We analyze the performance of the detector with different parameters - number of shots per class, and time limits for quick deployment. 

\begin{table}[t]
	\centering
% 	\setlength{\tabcolsep}{2mm}
% 	\fontsize{8}{9}\selectfont
% 	\begin{threeparttable}
	\caption{Performance of AirInteraction and TFA on all datasets. Bold values denote the best performance in each number of shot.}
	\label{tab:performance}
% 	\vspace{-0.1cm}
	\begin{tabular}{c|c|c|c|c}
		\toprule%[1.2pt]
% 		\multicolumn{1}{c}{} & \multicolumn{1}{c|}{} &
% 		\multicolumn{1}{c|}{1 shot} & \multicolumn{1}{c|}{2 shots} & \multicolumn{1}{c}{3 shots} \\
% 		\midrules
		Dataset & Model & 1 Shot & 2 Shots & 3 Shots\\
		\midrule
		SubT Tunnel & Ours ($r=1$) &  5.21 & 10.55 & 13.65 \\
		& Ours ($r=2$) & 12.39  & 14.78 & 14.29 \\
		& Ours ($r=3$) & \textbf{13.38} & \textbf{17.56} & \textbf{17.81} \\
		& TFA \cite{TFA} & 7.52 & 13.85 & 17.20\\
		\midrule
		SubT Urban & Ours ($r=1$) &  4.35 & 8.27 & 9.52 \\
		& Ours ($r=2$) & 8.32  & 10.20 & 11.51 \\
		& Ours ($r=3$) & \textbf{12.53} & \textbf{13.68} & \textbf{14.95} \\
		& TFA \cite{TFA} & 8.54 & 10.56 & 10.82\\
		\midrule
		Scott Reef 25 & Ours ($r=1$) & 8.37 & 10.02 & 34.72 \\
		& Ours ($r=2$) & \textbf{35.28} & 31.77 & \textbf{42.49} \\
		& Ours ($r=3$) & 32.48 & \textbf{34.62} & 37.74 \\
		& TFA \cite{TFA} & 28.25 & 30.92 & 36.97 \\
		\midrule
		Drone Filming & Ours ($r=1$) & 12.60 & 63.52 & 51.82 \\
		& Ours ($r=2$) & 40.03 & 83.41 & 83.25 \\
		& Ours ($r=3$) & \textbf{62.68} & \textbf{89.36} & \textbf{84.38} \\
		& TFA \cite{TFA} & 35.26 & 70.59 & 80.95\\
 		\bottomrule%[1.2pt]
	\end{tabular}
% 	\end{threeparttable}
\end{table}%

\subsection{Performance with Fixed Number of Shots}
\label{sec:performance-shots}

Since the online annotation is costly, we assume that human operator can give annotation feedback for up to 3 scenes per class. Therefore, we take very challenging settings to demonstrate the effectiveness of our method, i.e., only $k = 1,~2,~3$ shots are provided in the experiment. In addition, we also report the performance using the weighted mixture of batch for fine-tuning as described in \sref{sec:weighted}, i.e., the novel-to-base image ratio $r=1,~2,~3$ are used. 

% In this work, we differentiated the number of shots required to train each object category in few-shot object detection. 
% Contrary to normal few-shot learning work that evaluates various number of shots per class, we constrained the number of shots within 1, 2, 3, because we require human operator at the base station provides bounding box annotations of the objects real-time, and the input from the human operator is expensive. 
% We evaluate the performance of the AirInteraction in multiple perspectives. 
% First, we report the object detection performance of fine-tuned model, where novel shots are from the interaction with human annotator. 
In the \tref{tab:performance}, we reported the overall performance of AirInteraction, which is fine-tuned for 5 minutes to meet the quick deployment requirements.
% , across different number of shots (1, 2, 3) from human annotator, and across different novel-to-base image ratio $r=1,~2,~3$ in the minibatch.
The results show that AirInteraction with $r=2$ and $3$ achieve better performances than the TFA baseline in all three datasets, while AirInteraction with $r=1$ achieves lower performance than TFA. 

In average, the AirInteraction with $r=2$ and $r=3$ show \textbf{10.8\%} and \textbf{29.7\%} higher performance than the TFA baseline across all datasets. AirInteraction with $r=1$ shows 34.4\% lower performance than the TFA on average. This highlights the effect of tuning the ratio $r$ when employing AirInteraction; reusing the novel shots provided by the human operator in the minibatches facilitates the improved performance to the original model. We also note that the performance of AirInteraction with $r=1$ is relatively lower than the TFA baseline, because the dynamic addition of human-informed shots during the training is not as stable as starting the training with novel shots in the first place. Increasing the ratio $r$ can be thought of as a method to make up this shortcoming by using novel shots more frequently for training. 

In \tref{tab:class-performance}, we additionally reported the best performance of the AirInteraction model by object classes in the dataset. We observe that the AirInteraction is achieving comparatively good performance in detecting objects in Drone Filming and Scott Reef 25 datasets than SubT tunnel and SubT Urban datasets.

\subsection{Performance with a Fixed Training Time}

\begin{figure*}[!t]
    \centering
    \includegraphics[width=1.0\linewidth]{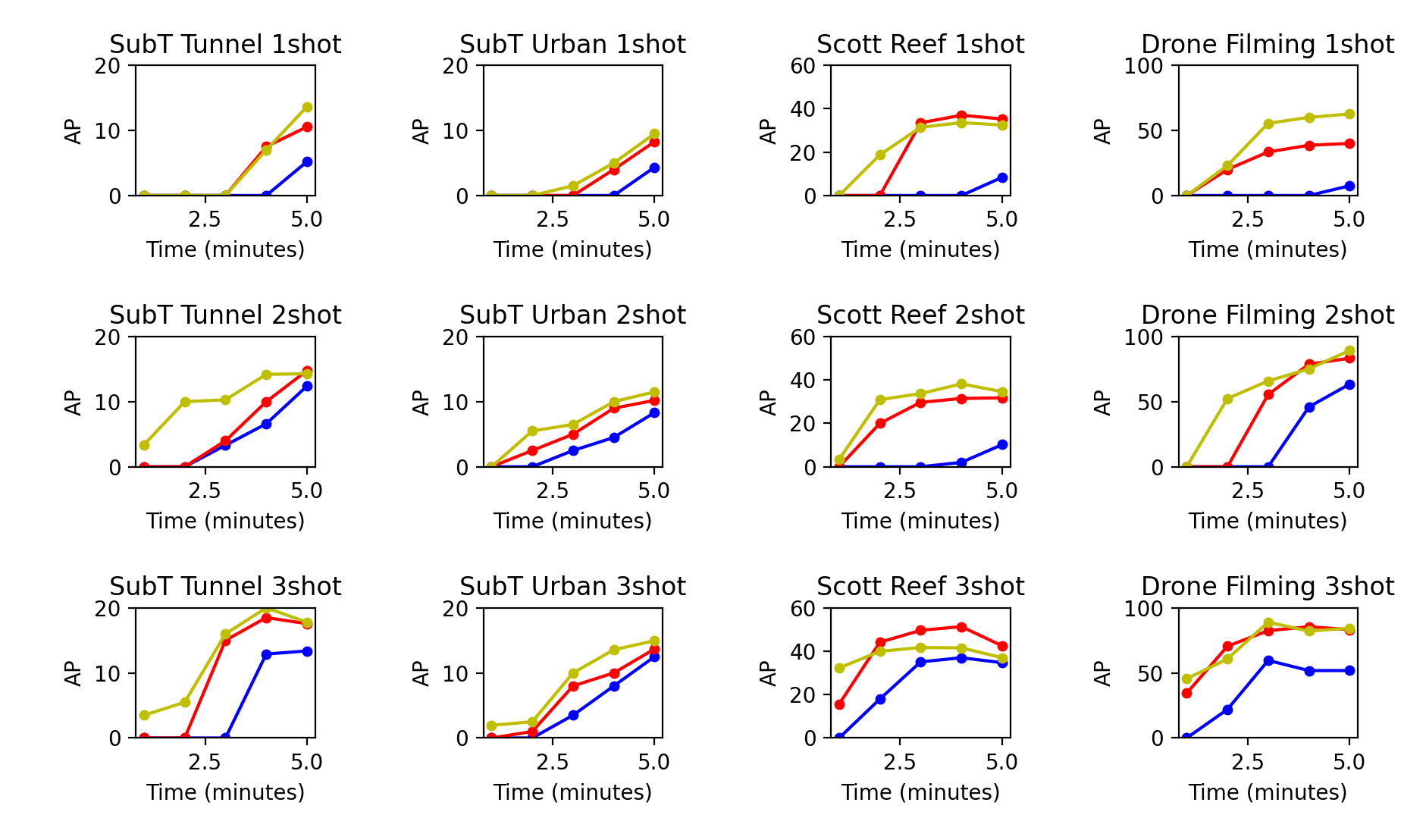}
    \caption{Performance of few-shot object detector within five minutes. Blue, red, yellow graph each denotes AirInteraction $r=1,2,3$}
    \label{fig:time}
\end{figure*}

In \sref{sec:performance-shots}, we report the model performance with fixed fine-tuning time.
However, during autonomous exploration, the robots should be able to learn the novel object categories within a limited amount of time for quick deployment.
In this section, we further report the model performance in \fref{fig:time} with a fine-tuning time of 1, 2, 3, 4, and 5 minutes. 
% measure the training time of the few-shot object detector, and evaluate the performance of the few-shot object detector at each time checkpoints. 
This means that the few-shot object detector should learn novel object categories within five minutes for exploration, which is acceptable for most tasks. In general, the performance of the object detector improved over time. In the case of $r=2$ and $3$, the performance of the object detector started to saturate starting 3 minutes, while in the case of $r=1$, the performance of object detector was low until 5 minutes. On average, across the datasets, the model performance of AirInteraction with $r=2$ at time $t=3$ and $4$ minutes was $80.8\%$ and $97.5\%$ of its performance at $t=5$ minutes, respectively, and the performance of AirInteraction $r=3$ at time $t=3$ and $4$ minutes was $85.6\%$ and $94.8\%$ of its performance at $t=5$ minutes.

\begin{table}[t]
  \centering
%   \setlength{\tabcolsep}{2.1mm}
%   \fontsize{9}{10}\selectfont
  \caption{Best Object detection performance of AirInteraction for each class of object in each dataset.}
  \label{tab:class-performance}
    \begin{tabular}{c|cc}
    \toprule%[1.2pt]
    Dataset & Class & Performance\\
    \midrule
    SubT Tunnel & Door &  19.28\\
    & Hole &  10.25\\
    & Cctv &  14.55\\
    & Fire Extinguisher &  18.50 \\
    \midrule
    SubT Urban & Backpack &  14.38 \\
    & Helmet &  18.25 \\
    & Rope  & 9.44 \\
    & Drill & 11.69 \\
    & Vent  &  16.28 \\
    & Survivor & 13.74 \\
    \midrule
    Scott Reef 25 & Weed & 32.54 \\
    & Finger Coral & 53.15 \\
    & Mushroom Coral & 43.16 \\
    \midrule
    Drone Filming & Van & 98.52\\
    & Truck & 83.41\\
    & Bicyler & 88.11\\
    \bottomrule%[1.2pt]
    \end{tabular}%
  \label{tab:subt_cls}%
\end{table}

\subsection{Online Responsivity}

Real-time performance is crucial for robotic systems, thus we next measure the online responsivity of the interestingness detection system. 
Specifically, we employ the metric Area Under Curve of Online Precision (AUC-OP) \cite{eccv}, which jointly considers online response, precision, and recall rate.
% We compared the performance of AirInteraction's online interestingness responses. 
The baseline is the online visual interestingness algorithm \cite{eccv} without the object detector component, which detects the binary interestingness of the scenes only using a visual memory module. Basically, all the scenes that contain interesting objects are taken as interesting, and the others are not interesting. 
% we measured the online responses of AirInteraction by the same metric, AUC-OP. 

In the \tref{tab:auc-op}, we reported the AUC-OP scores of the baseline visual interestingness algorithm and AirInteraction in the different datasets. In the SubT Tunnel, SubT Urban, Scott Reef 25, and Drone Filming datasets, the AirInteraction showed better online response scores than the baseline.
Note that we often accept several false positives to improve the recall rate and the metric of AUC-OP provides a variable $\delta~(\delta\ge1)$ to reflect this. Basically, higher $\delta$ means more false positives are allowed and $1$ means no false positive is allowed ($\delta=2$ is recommended for most applications).

\begin{table}[h]
  \centering
%   \setlength{\tabcolsep}{2.1mm}
%   \fontsize{9}{10}\selectfont
  \caption{The performance on AUC-OP score.}
    \begin{tabular}{c|c|ccc}
    \toprule%[1.2pt]
    Dataset & Method & $\delta=1$ & $\delta=2$ & $\delta=4$\\
    \midrule
    SubT Tunnel & Baseline \cite{eccv} &  0.32 & 0.51 & 0.72\\
    & Ours & 0.36 & 0.53 & 0.81 \\
    \midrule
    SubT Urban & Baseline \cite{eccv}&  0.29 & 0.45 & 0.66 \\
    & Ours & 0.35 & 0.44 & 0.69 \\
    \midrule
    Scott Reef 25 & Baseline \cite{eccv} &  0.44 & 0.71 & 0.89\\
    & Ours & 0.58 & 0.75 & 0.85 \\
    \midrule
    Drone Filming & Baseline \cite{eccv} &  0.56 & 0.68 & 0.74\\
    & Ours & 0.59 & 0.70 & 0.88\\
    \bottomrule%[1.2pt]
    \end{tabular}%
  \label{tab:auc-op}%
\end{table}

\subsection{Reducing Communication Bandwidth}

The purpose of online unsupervised learning module is to select a few interesting scenes and send those scenes to the base station, thereby reducing communication bandwidth between the robot system and the human operator. In this section, we measure the reduction of communication bandwidth by comparing the ratio of the number of selected interesting scenes sent to the base station and the number of total scenes. As can be seen in \tref{tab:communication_bandwidth}, our system reduce the communication bandwidth 80 \% up to 91\% across datasets, which indicates its effectiveness.

\begin{table}[h]
  \centering
%   \setlength{\tabcolsep}{2.1mm}
%   \fontsize{9}{10}\selectfont
  \caption{Percentage of images being sent}
    \begin{tabular}{c|cccc}
    \toprule%[1.2pt]
     & SubT-Tunnel & SubT-Urban & SR25 & DF\\
    \midrule
    Ratio & 0.15 & 0.11 & 0.20 & 0.09 \\
    \bottomrule%[1.2pt]
    \end{tabular}%
  \label{tab:communication_bandwidth}%
\end{table}

\section{Limitation} \label{sec:limitation}
One of the limitations of AirInteraction comes from the fact that it is using a fine-tuning based few-shot object detector. Under our specific choice of fine-tuning-based object detection model \cite{TFA}, the maximum number of novel object classes should be fixed; we set the capacity of new object categories as 10, but if the robot faces more objects during the mission, the capacity should be enlarged. Moreover, fine-tuning the model requires some amount of time to reach a good performance; although we limited the training time to five minutes, and showed that AirInteraction achieves good performance within this time constraint, it is better to shorten the amount of time learning a new category. To this end, deploying a few-shot object detection model without fine-tuning should be considered. Recent works on dynamic few-shot object detection without fine-tuning \cite{airdet} or incremental meta-learning based few-shot object detection \cite{ONCE} can be our future alternatives.

\section{Conclusion}
We proposed the human-robot interactive framework, AirInteraction, that enables interesting scene recognition via human-informed few-shot object detection. Composed of three components, (1) online unsupervised learning module, (2) human-robot interaction interface, and (3) few-shot object detector, AirInteraction aids mobile unmanned exploration robots to learn novel interesting objects during the mission only relying on minimum amount of human feedback. Our additional techniques, weighted mixture minibatches and parameter synchronization, improve the performance of the few-shot object detector in the online learning setting that requires quick responses. Our experiment results show that AirInteraction achieves good object detection performance in diverse object detection datasets, and impressive performance within less than five minutes. 


%%%%%%%%%%%%%%%%%%%%%%%%%%%%%%%%%%%%%%%%%%%%%%%%%%%%%%%%%%%%%%%%%%%%%%%%%%%%%%%%

%%%%%%%%%%%%%%%%%%%%%%%%%%%%%%%%%%%%%%%%%%%%%%%%%%%%%%%%%%%%%%%%%%%%%%%%%%%%%%%%

%%%%%%%%%%%%%%%%%%%%%%%%%%%%%%%%%%%%%%%%%%%%%%%%%%%%%%%%%%%%%%%%%%%%%%%%%%%%%%%%

\section*{Acknowledgment}
This work was partially sponsored by the ONR grant \#N0014-19-1-2266 and ARL DCIST CRA award W911NF-17-2-0181.
The authors would like to thank all members of the Team Explorer for providing data collected from the
DARPA Subterranean Challenge.

%%%%%%%%%%%%%%%%%%%%%%%%%%%%%%%%%%%%%%%%%%%%%%%%%%%%%%%%%%%%%%%%%%%%%%%%%%%%%%%%

\bibliographystyle{IEEEtran}
\bibliography{IEEEabrv, IEEEexample}

\end{document}